\documentclass[sigconf]{acmart}

\usepackage{mdframed}
\usepackage{amsmath}
\usepackage{booktabs}
\usepackage{multirow}
\usepackage{adjustbox}
\usepackage{algorithm}
\usepackage{algpseudocode}
\usepackage{amsmath}
\usepackage{xcolor}

\newcommand{\SA}[1]{\textcolor{black}{#1}}
\usepackage{tikz}
\usepackage{tcolorbox}
\newcommand\mybox[2][]{\tikz[overlay]\node[fill=blue!20,inner sep=2pt, anchor=text, rectangle, rounded corners=1mm,#1] {#2};\phantom{#2}}
\newcommand{\gbox}[1]{\mybox[fill=green!20]{#1}}

\newcommand{\tikzxmark}{%
  \tikz[x=1.5ex, y=1.5ex, line width=.5ex, red]%
    \draw (0,0) -- (1,1) (1,0) -- (0,1);%
}
\makeatletter
\renewcommand{\ALG@name}{Algorithm}

\algrenewcommand\algorithmicindent{1.5em}
\makeatother

\newcommand{\para}[1]{\paragraph{\textnormal{\textbf{#1}.}}} 
\usepackage{algorithm} 
\usepackage[T1]{fontenc}

\usepackage[utf8]{inputenc}

\usepackage{microtype}

\usepackage{graphicx}

\usepackage{pifont}
\begin{document}

\title{\texttt{PROSLEX}: A Novel Dataset for Expert-Annotated Legal Statute Prediction for Indian Judiciary}

%
\author{Subinay Adhikary}
\email{sa21rs094@iiserkol.ac.in}
\affiliation{%
  \institution{IISER Kolkata}
  \country{India}
}

\author{Upal Bhattacharya}
\email{upal.bhattacharya@gmail.com}
\affiliation{%
  \institution{Utrecht University}
 \country{Netherlands}
  }

\author{Vivek Kumar Singh}
\email{Vksingh306@gmail.com}
\affiliation{%
  \institution{IISER Kolkata}
  \country{India}
}

\author{Anurag Sharma}
\email{anuragsharma3211@gmail.com}
\affiliation{%
 \institution{IIT Kharagpur}
 \country{India}
 }

\author{Shubham Kumar Nigam}
\email{shubhamkumarnigam@gmail.com}
\affiliation{%
  \institution{University of Birmingham Dubai}
  \country{United Arab Emirates}
}

\author{Suvasis Das}
\email{suvasisdas7@gmail.com}
\affiliation{%
  \institution{IISER Kolkata}
  \country{India}
  }

\author{Shouvik Kumar Guha}
\email{shouvikkumarguha@nujs.edu}
\affiliation{%
  \institution{WBNUJS}
  \country{India}
  }

\author{Koustav Rudra}
\email{krudra@cai.iitkgp.ac.in}
\affiliation{%
  \institution{IIT Kharagpur}
  \country{India}
  }

\author{Kripabandhu Ghosh}
\email{kripa.ghosh@gmail.com}
\affiliation{%
  \institution{IISER Kolkata}
  \country{India}
}

%
\renewcommand{\shortauthors}{Adhikary et al.}

\begin{abstract}
  Legal Statute Prediction (LSP) involves automatically identifying relevant legal statutes given factual descriptions in legal documents, typically framed as a \textit{multi-label classification task} within natural language processing and information retrieval research. While recent advances have begun incorporating Large Language Models (LLMs) for statute prediction, current approaches primarily focus on accuracy metrics without addressing the critical need for legal reasoning—a fundamental requirement in judicial contexts where decisions must be explainable and justifiable. To address this research gap, we present \texttt{PROSLEX} ({\bf PR}ediction {\bf O}f {\bf S}tatutes and {\bf LE}gal e{\bf X}planation), a comprehensive dataset comprising \textbf{1,623 expert-annotated} legal documents from the Indian context. Each document is paired with statute predictions and detailed explanations, \textbf{totaling 7,450 explanations}, capturing the underlying legal reasoning. Using this dataset, we systematically evaluate various prompting strategies, including zero-shot, few-shot, chain-of-thought, and tree-of-thoughts approaches, to generate both statute predictions and their corresponding legal rationales. Our evaluation framework measures not only predictive performance but also the coherence and legal validity of generated explanations, positioning PROSLEX as a benchmark for developing explainable AI systems that can support legal practitioners while advancing research in interpretable legal NLP. To ensure reproducibility, we have made our \emph{PROSLEX} dataset and model code available on GitHub\footnote{\url{https://github.com/subinay494/Legal_Statute_Prediction_Explanation}}.
\end{abstract}



\keywords{Statute Prediction, Large Language Model, Explanation}


\acmConference[ICAIL 2026]{International Conference on Artificial Intelligence and Law}{June 2026}{Singapore}
\acmYear{2026}
\copyrightyear{2026}
\maketitle

\section{Introduction}
In Civil Law jurisdictions, written statutory provisions explicitly define \texttt{charges} under specific \texttt{statutes} (e.g., \emph{theft}, \emph{murder}, \emph{riot}), which must be consulted before determining the applicable charge in a given case\footnote{\url{https://en.wikipedia.org/wiki/Civil_law_(legal_system)}}. `Statute' refers to a set of codified legal rules followed by the legal system in the concerned judiciary. In countries such as India, which adhere to the Civil Law system (in conjunction with elements of Common Law) and face the looming pendency of millions of cases\footnote{\url{https://njdg.ecourts.gov.in/njdg_v3/}}, automating statute identification emerges as a significant research challenge -- not only to support legal professionals but also to aid individuals without legal expertise, for whom seeking professional advice can be costly and time-consuming. Statute prediction makes it easy to find the laws that apply to a case, giving both lawyers and everyday people quick, affordable access to legal guidance.

 \begin{figure*}[t]
\centering
\begin{tcolorbox}[title={An example document annotated by legal experts.}]
\small
  BHIMANNA v. STATE OF KARNATAKA [2012] INSC 495 (4 September 2012)  On 17.11.1999 at about 4.00 p.m., Yenkappa(A-1), along with Bhimanna (A-2) and Suganna (A-3), was returning home with agricultural implements, that is axes and a plough.  They attempted to use the land of the deceased as a pathway.  The deceased Bheemanna, who was present on his land along with his wife Paddamma (PW.1) and mother, namely, Bheemava, obstructed the accused persons, asking them not to pass through his land.  Yenkappa(A-1) then started hurling $\underbrace{\text{\gbox{abuses in filthy language and instigated Bhimanna (A-2) and Suganna (A-3) to assault the deceased.}}}_{Statute:  \text{\textbf{IPC 147}}}$  $\underbrace{\text{\gbox{Thus, Bhimanna (A-2) and Suganna (A-3) began assaulting the deceased with axes over his head and right hand.}}}_{Statute: \text{\textbf{IPC 302, IPC 147}}}$ Yenkappa (A-1) assaulted the deceased with the wooden part of a plough.  Paddamma (PW.1) and Bheemava, mother of the deceased went to save the deceased, but they too, were threatened with assault.  Similar threats were hurled when Rangayya (PW.6), nephew of the deceased and his father Hanumappa approached the place of occurrence.  The accused persons left the place after assaulting the deceased, throwing away the axes and wooden part of the plough.  Rangayya (PW.6) brought a bullock cart as asked by Paddamma (PW.1) from the village and the deceased was then taken to Ramdurga Police Station.  Upon the advice of the police, the deceased was taken in a mini lorry, driven by Mahadevappa (PW.10) to Deodurga Hospital, and when they reached there at 8.00 p.m., $\underbrace{\text{\gbox{based on the complaint submitted by Paddamma (PW.1), an FIR was lodged at 8.15 p.m.}}}_{Statute: \text{\textbf{IPC 302}}}$
\end{tcolorbox}
\caption{An example of annotation by legal experts in the proposed \texttt{PROSLEX} dataset. Here, statutes IPC 147 and IPC 302 are applicable, and the explanation (reason) for the applicability of these statutes is shown as {\color{green} annotations}.}
\label{fig:example_annotation}
\end{figure*}

\textit{Statute prediction} involves identifying the relevant legal statutes for a given case based on its factual description, making it a practical and impactful application of Natural Language Processing~\cite{paul2024legal,chalkidis2021lexglue, adhikary2024case}.
Since a legal case document can be associated with multiple statutes, statute prediction is inherently a \emph{multi-label classification} task. 

Explainability in AI-systems, especially in high-risk domains like law, is of paramount importance \cite{Liao2024AI}. The EU AI Act Article 13: Transparency and Provision of Information to Deployers \footnote{\url{https://artificialintelligenceact.eu/article/13/}} states {\it ``{\bf High-risk AI systems} shall be designed and developed in such a way as to ensure that their operation is {\bf sufficiently transparent to enable deployers to interpret a system’s output} and use it appropriately. An appropriate type and degree of transparency shall be ensured with a view to achieving compliance with the relevant obligations of the provider and deployer set out in Section 3''} highlighting the non-negotiable necessity of transparency in AI systems. To build explainable automatic statute prediction systems, test beds are necessary, and in Indian law, apart from initial attempts like \cite{vats-etal-2023-llms}, there has been no large-scale court facts (of cases) dataset containing {\it legal expert annotated} explanations for the applicability of statute(s). In this paper, we present \texttt{PROSLEX} ({\bf PR}ediction {\bf O}f {\bf S}tatutes and {\bf LE}gal e{\bf X}planation), a dataset comprising {\bf 1,623} legal expert–annotated case facts paired with their relevant statutes for explainable statute prediction.  Figure~\ref{fig:example_annotation} illustrates an example annotation, where legal experts highlighted a text span and associated it with the applicable statute. Afterward, \textbf{a total of 7,450 spans} were annotated to represent the legal reasoning for the statutes.

\begin{figure*}
    \centering
    \includegraphics[width=0.8\textwidth]{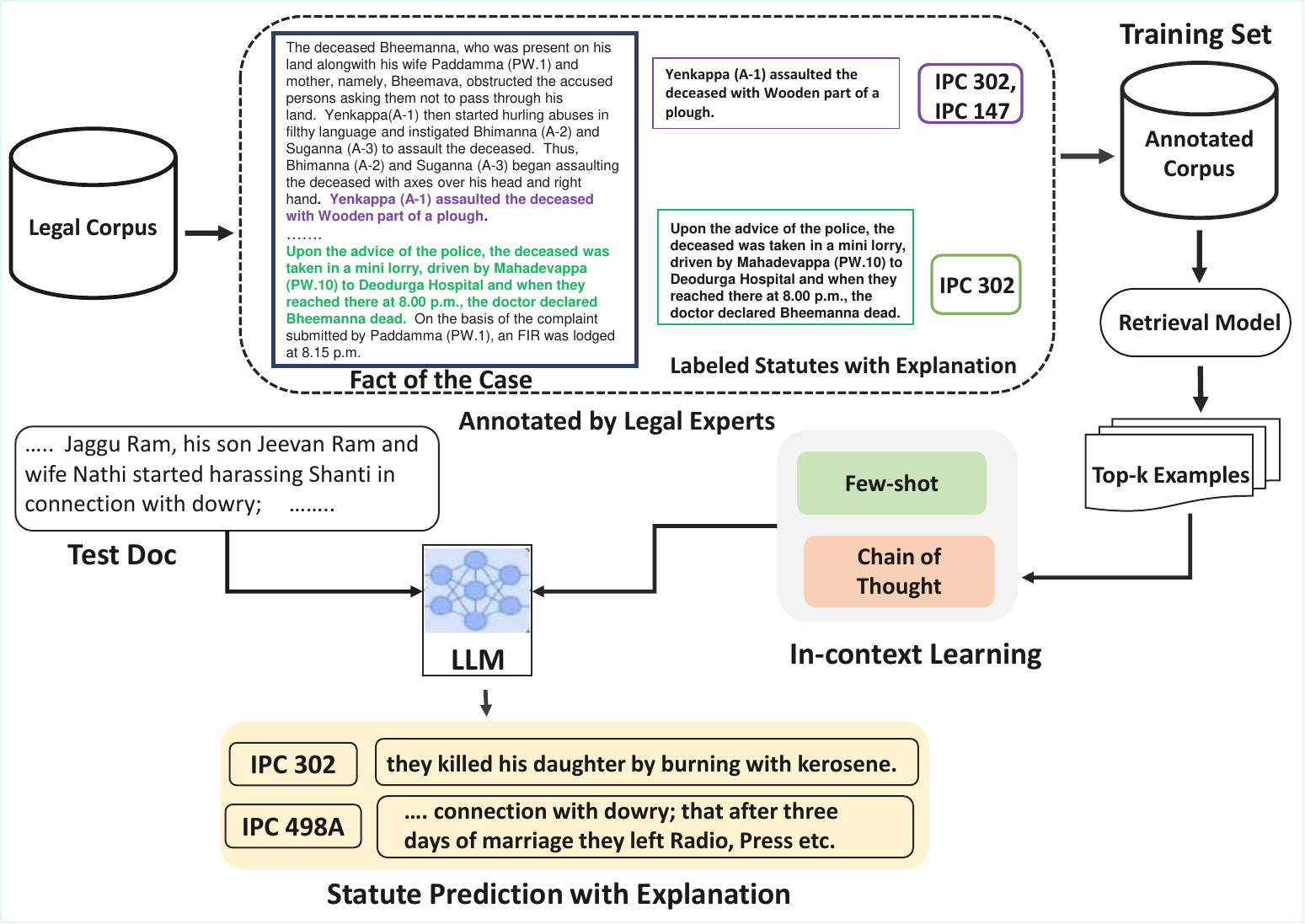}
    \caption{This workflow illustrates a legal prediction framework to identify applicable Indian Penal Code (IPC) statutes for legal cases with the reasoning. The diagram shows a workflow where a legal corpus is annotated by legal experts to create a training set of case facts paired with relevant statutes and explanations. The left side displays an example case involving Bheemanna's assault and death, with highlighted text mapped to IPC sections 302 (\emph{murder}) and 147 (\emph{rioting}). The system uses a retrieval model to find similar cases, then employs in-context learning techniques like few-shot examples and chain-of-thought reasoning to make predictions on unlabeled cases. The bottom portion demonstrates the model analyzing a test document about dowry harassment and correctly predicting applicable statutes (IPC 302 for a \emph{kerosene burning death} and IPC 498A for \emph{dowry harassment}).}
    \label{fig:workflow}
\end{figure*}

The creation of \texttt{PROSLEX} involved two main stages and was carried out by legal experts from a reputed Indian legal institute. In the first stage, experts \textit{identified statutes} based on their legal and social importance. In the second stage, they \textit{selected documents} corresponding to these statutes and annotated the relevant text spans. Subsequently, a senior legal expert resolved annotations with low inter-annotator agreement (IAA) scores. Section~\ref{sec:dataset} provides a detailed description of the \texttt{PROSLEX} creation process. The resulting dataset was then used for our empirical study, which is elaborately described in Section~\ref{sec:experiment}. Furthermore, Figure~\ref{fig:workflow} depicts the complete pipeline, from dataset creation to its application in downstream tasks using various in-context learning techniques, including \textit{few-shot}, \textit{chain-of-thought}, and \textit{tree-of-thoughts}.

Our research delivers a comprehensive AI-based system for legal statute prediction and explanation tailored to the Indian judiciary. 

\begin{table*}[t]
    \centering
    \caption{A Comparative Overview of Prominent Legal Statute Prediction Datasets.}
   \resizebox{0.95\textwidth}{!}{\begin{tabular}{||c c c c c c ||} 
 \hline
 \textbf{Corpus} & \textbf{Language}  & \textbf{Jurisdiction} & \textbf{\# of Cases} &  \textbf{\# Labels} & \textbf{Explanation} \\ [0.5ex] 
 \hline\hline
 ECtHR-B~\cite{chalkidis2021lexglue} & English & Europe & 10,000 & 10 & \tikzxmark  \\
 \hline
 ILSI~\cite{paul2024legal} & English & India &  66,090 & Law Article (100) & \tikzxmark \\ 
 \hline
 LeSICiN~\cite{paul2022lesicin} &English &  India & 100,000  & IPC Statutes (multiple)& \tikzxmark \\ 
 \hline
  FIRE AILA 2019~\cite{bhattacharya2019overview} & English & India & 100 & 197 & \tikzxmark \\
  \hline 
  \cite{vats-etal-2023-llms}& English & India & 45 & 80 & {\color{green}\ding{52}} \\
  \hline \hline
 \textbf{PROSLEX} (Ours) & English & India & 1623 
  & 7 & {\color{green}\ding{52}}   \\  [1ex] 
 \hline
\end{tabular}}
    
    \label{tab:dataset}
    
\end{table*}
\section{Related Work}
Legal charge identification or statute prediction has significantly evolved, leveraging advances in machine learning and deep learning methodologies. Early research primarily relied on traditional machine learning methods. For instance, \citet{LIU2015194} utilized support vector machines (SVMs) with textual features from legal documents to predict associated statutes.

Motivated by the success of deep learning in NLP tasks~\cite{bambroo2025marro,adhikary2025thinkir}, researchers introduced neural network architectures into charge prediction tasks~\cite{chen2019charge, li2018law, shen2018legal, undavia2018comparative, wei2018empirical, yang2019recurrent}. \citet{luo2017learning} proposed a hierarchical attention network that predicted charges by selecting the most relevant law articles. Similarly, \citet{wei2019external} developed a knowledge-aware multi-label charge prediction method using external knowledge from legal texts to improve prediction accuracy.

On the other side, to address the challenges of distinguishing confusing charges, \citet{li2019element} developed an element-driven attentive neural network model leveraging legal constitutive elements as discriminative features. Similarly, \citet{yang2019legal} integrated word collocation features using attention mechanisms, and \citet{hu2018few} constructed discriminative attributes to enhance differentiation between confusing charges. More recently, \citet{xu2020distinguish} developed an attention mechanism focusing on subtle differences between confusing legal articles, aiming to improve charge prediction accuracy further.

Knowledge-enhanced models have shown effectiveness, particularly for confusing charges. \citet{bi2024knowledge} presented a dual-graph interaction framework that integrates external legal knowledge bases, significantly improving performance in distinguishing among confusing charges. Moreover, \citet{chen2024learning} introduced a defendant-aware label representation method tailored for multi-defendant scenarios, addressing complexities in real-world legal cases. Several works emphasized understanding semantic relations between charges and textual descriptions to enhance prediction accuracy~\cite{xu2020distinguish,yue2021neurjudge,li2023leveraging,kang2021label}. 

Within the Indian legal context, the Artificial Intelligence for Legal Assistance (AILA) track at FIRE 2019 was instrumental in emphasizing legal information retrieval and statute identification~\cite{bhattacharya2019fire}. Subsequently, \citet{vats-etal-2023-llms} investigated the strengths and limitations of large language models (e.g., GPT, Llama, Gemini) for statute prediction, demonstrating both substantial promise and critical shortcomings. Further, comprehensive state-of-the-art surveys by \citet{pawar2023extraction} and \citet{paul2024legal} on publicly available datasets underscored notable progress while also identifying persistent challenges in Indian legal statute prediction.


Despite these advances, a notable gap persists in \textit{generating clear explanations alongside charge predictions}, especially within the in-context learning framework. In response to this limitation, we introduce \texttt{PROSLEX}, a novel dataset created to enable in-depth analysis of charge identification with accompanying explanatory context, focusing on Indian legal cases.

\section{Dataset Creation}\label{sec:dataset}
In this section, we provide a detailed account of the creation process of our dataset, \textbf{PROSLEX} ({\bf PR}ediction {\bf O}f {\bf S}tatutes and {\bf LE}gal e{\bf X}planation), which comprises a subset of 33,546 Supreme Court of India\footnote{https://www.sci.gov.in/} judgements from 1950 to 2016, covering 444 Acts and 3,428 Sections.  

\subsection*{Annotation Process}
\subsection{Involvement of Legal Experts}

The dataset creation was supported by a \textbf{Government-funded initiative, involving two postgraduate legal experts per statute, and a senior faculty member from a reputed National Law University}. Legal experts initiate the annotation process by first selecting \textit{relevant statutes} and then choosing the \textit{corresponding documents} for each statute.

\begin{table*}[t]
\footnotesize
\caption{\textbf{Statistics of the \texttt{PROSLEX} Dataset}}
\resizebox{1\textwidth}{!}{
    \begin{tabular}{|l|c|c|c|c|c|c|c|c|}
    \hline
 \multicolumn{2}{|c|}{\textbf{Dataset}} &\multicolumn{7}{|c|}{\textbf{Indian Penal Code (IPC) Statutes} }\\
    \hline
Type & Statistics & IPC 147 & IPC 201 & IPC 302 & IPC 376 & IPC 420 & IPC 498A & IPC 506 \\
     \hline
    Train  & \#Documents & 214 & 214 & 193 & 105 & 172 & 118 & 116  \\
      \hline
    Train  & \# Avg. tokens & 1052 & 1584 & 617 & 745 & 496 & 270 & 488 \\
     \hline
   Validation & \#Documents & 30 & 30 & 27 & 15 & 24 & 17 & 16\\
    \hline
   Validation & \# Avg. tokens  & 1100 & 1540 & 583 & 750 & 487 & 274 & 498\\
    \hline
    \hline
     Test  & \#Documents & 62 & 63 & 57 & 30 & 51 & 35 & 34 \\
       \hline
     Test  & \# Avg. tokens & 1043 & 1534 & 600 & 723 & 415 & 323 & 502 \\
    \hline
\end{tabular}}


\label{tab:dataset_stats}

 \end{table*}

. \begin{table}[h]
\centering
\caption{The structure of the prompt used in our ICL experiments. [X] represents a variable that is to be substituted for its value.}
\small
\begin{tcolorbox}[
 colback=blue!5!white,
    colframe=blue!75!black,   
  title=Prompt Template for Predicting Statute with Explanation, 
  coltitle=white,            
  colbacktitle=blue!70!black,
  fonttitle=\bfseries,
  boxrule=1.2pt,
  arc=3pt,
  left=3mm,
  right=3mm,
  top=1mm,
  bottom=1mm
]

Given a fact of a legal case document as enclosed within angular brackets, enlist the likely statutes that apply to this segment.\\

Following are the tags along with their descriptions.\\
1. ``[Statute]'': ``[Description]'' \\  
2. ``[Statute]'': ``[Description]''  \\
... \\

Following is a list of example text for each :\\

\# Example 1:\\
 Case\_proceeding: ``[Text]''\\
 Prediction: ``[Statute],[Statute]''\\
 Explanation: ``[Reasoning],[Reasoning]''\\

 \# Example 2:\\
 Case\_proceeding: ``[Text]''\\
 Prediction: ``[Statute]''\\
 Explanation: ``[Reasoning]''\\

Instruction: Learn from the examples provided. Avoid generating fabricated or invalid tags.\\

Input segment: <[INPUT]>
\end{tcolorbox}

\label{tab:prompt_template}
\end{table}

\subsection{Statute selection} We concentrate on crime-related documents, which account for 30$\%$ of the Supreme Court collection. In the Indian judiciary, 512 sections related to criminal law, collectively known as the Indian Penal Code (IPC). The selection of statutes for annotation was guided by the following three criteria:

\begin{enumerate}
    \item \textbf{Legal and social significance:} Those statutes have high importance from a legal perspective and the socio-legal perspective. For example, Indian Penal Code 302 represents the sentence for murder and is thus of high relevance. 
    \item \textbf{Sufficient documents:} Those statutes have a substantial representative number of documents for the empirical study.
    
    \item \textbf{Distinctiveness of legal provisions:}  The nature of statutes needs to be distinct. For instance, IPC 302 indicates a sentence (punishment) for culpable homicide amounting to murder, IPC 300 defines culpable homicide, and IPC 304 addresses culpable homicide but not amounting to murder. Hence, to select the distinctive statute, legal experts only select those documents that are associated with IPC 302 as it concerns the sentence for the most serious crime within {\it homicide}.\footnote{\url{https://devgan.in/ipc/chapter_16.php}}
\end{enumerate}

\para{Statute overview}\label{sec:statute_detail} Based on the aforementioned criteria, the legal experts identified \textbf{seven statutes} that meet our requirements, which are described below:

\begin{enumerate}
    \item \textbf{Indian Penal Code, $1860\_147$}: 
    Whoever is guilty of \emph{rioting}, shall be punished with imprisonment of either description for a term which may extend to two years, or with fine, or with both.
    \item \textbf{Indian Penal Code, $1894\_201$}:
    Whoever, knowing or having reason to believe that an offence has been committed, \emph{causes any evidence of the commission of that offence to disappear}.
    \item \textbf{Indian Penal Code, $1860\_302$}:
    Whoever commits \emph{murder} shall be punished with death, or imprisonment for life, and shall also be liable to fine.
    \item \textbf{Indian Penal Code, $1860\_376$}: Deals with the \emph{punishment for rape}.
    \item \textbf{Indian Penal Code, $1860\_420$}:
    Whoever cheats and thereby dishonestly induces the person deceived to deliver any property to any person.
    \item \textbf{Indian Penal Code, $1860\_498A$}:
    An act of \emph{cruelty} for the purposes of Section 498A corresponded to willful conduct of such a nature that it could cause danger to the life, limb, or health of the woman.
    \item \textbf{Indian Penal Code, $1860\_506$}:
    Whoever commits the \emph{offence of criminal intimidation} shall be punished with imprisonment of either description for a term which may extend to two years, or with fine, or with both.
\end{enumerate}

\subsection{Document Selection and Explanation Extraction}
Manually annotating all 33,546 documents is both time-consuming and labor-intensive, and expensive. As a solution, we opted to select a representative \textit{subset of collection} for annotation.
In this phase, legal experts focus on selecting a substantial amount of \textit{relevant documents} of the aforementioned statutes. Considering the \textbf{budget constraints} of the Government-funded project, the selection was restricted to at most $250$ cases per section.  To optimally utilize the available resources, we prioritized including a greater number of cases per statute, ensuring sufficient data points for each class, over covering a larger number of statutes with fewer cases. Now, legal experts focus on two key tasks: i) \textit{document selection} and ii) \textit{extracting text span (i.e., explanation) relates with corresponding statute}, which is carried out by a team of two legal experts per section, and is described below.

Let the section is  $S_k$ and the two legal experts are $L_1$ and $L_2$. 

\begin{enumerate}
    \item $L_1$ and $L_2$ individually selected cases (same cases annotated by both the annotators) in which \textit{section $S_k$ was upheld by the ruling court}. Thus, the cases in which section $S_k$ was initially applied but ultimately repealed are struck off from the list of cases to be annotated.
    \item The legal experts $L_1$ and $L_2$, extract \textit{facts of the case} from the original case document.
    \item The legal experts $L_1$ and $L_2$, perform annotation over facts of the cases. Figure~\ref{fig:example_annotation} displays legal experts labeling a \textit{span of text} along with applicable statutes, as the explanation.  
    \item Later, to create the ground-truth version, we follow an adjudication process with the help of a senior legal expert. 
\end{enumerate}

    \subsection{Adjudication Process}
\SA{To establish the gold standard annotations, we measured inter-annotator agreement between explanation-statute pairs provided by two independent legal experts. We computed text overlap using ROUGE-L~\cite{lin2004rouge}, which relies on the longest common subsequence to capture sentence-level structural similarity and word order. The scores range from 0 (no overlap) to 1 (perfect overlap). Higher scores indicate stronger consistency between annotators. Analysis of ROUGE-L scores across all annotation pairs revealed:}
\begin{itemize}
    \item Mean ROUGE-L: 0.79 (SD = 0.12)
    \item Median: 0.80
    \item Quartiles: Q1 = 0.72, Q3 = 0.87
\end{itemize}

\SA{To determine an appropriate threshold for requiring adjudication, we conducted a calibration study with 50 randomly sampled annotation pairs. A senior legal expert with over 15 years of experience in legal domain blind-reviewed each pair and indicated whether adjudication was necessary. This analysis showed that cases with ROUGE-L $<$ 0.75 required adjudication in 90\% of instances.}

Based on these findings, we adhere to the following adjudication steps:
\begin{itemize}
    \item \textbf{ROUGE-L $<$ 0.75:} The senior legal expert independently reviewed both annotations and either selected the more accurate one or created a refined version combining insights from both annotators.
    \item \textbf{ROUGE-L $\geq$ 0.75:} In this case, we retained the annotation with more detailed legal reasoning.
\end{itemize}

\SA{This approach ensured high-quality gold standard annotations while maintaining annotation efficiency, with approximately 37\% of cases requiring senior expert adjudication.}

\SA{The process of \textit{fact extraction}, \textit{labeling span of text}, \textit{solving inter-annotator disagreement}, and providing additional cases if required, are repeated until for section $S_k$ we have 250 relevant and annotated cases. Although it was not feasible to obtain $250$ cases for every statute, our dataset \texttt{PROSLEX} nevertheless offers \textit{1623 annotated documents with 7450 explanations}, based on Indian Supreme Court judgments (as shown in Table~\ref{tab:dataset}). We allocated \textit{94 INR per document, resulting in a total expenditure of approximately 152,562 INR for the complete annotation process}. This amount represented the entire budget for the project, which also limited our ability to annotate a larger number of cases. In Table~\ref{tab:dataset_stats}, we showcase overall statistics of our dataset \texttt{PROSLEX}. }
 Later, we employ \texttt{PROSLEX} for various empirical studies.







\begin{table*}[t]
\centering
\caption{Performance comparison of different models (e.g., DeepSeek, Llama-3.1-70B, Claude, and GPT-4) across sections of the Indian Penal Code. This evaluation presents the F1-scores of various statutes using both LM-based and LLM-based approaches. Among the LM-based models, InLegalBERT consistently outperforms others. Additionally, a key finding is that LLMs achieve the highest performance in statute prediction with explanation when using the Chain-of-Thought (CoT) setup within In-Context Learning (ICL) techniques.}
\renewcommand{\arraystretch}{1.2}

\begin{adjustbox}{max width=\textwidth}
\begin{tabular}{|l|l|c|c|c|c|c|c|c|c|}
\hline
\multirow{2}{*}{} & \multirow{2}{*}{\textbf{Models}} & \multicolumn{7}{c|}{\textbf{Indian Penal Code (Statutes)}} & \multirow{2}{*}{\textbf{Macro F1 score}} \\

\cline{3-9}
 & & \textbf{IPC 147} & \textbf{IPC 201} & \textbf{IPC 302} & \textbf{IPC 376} & \textbf{IPC 420} & \textbf{IPC 498A} & \textbf{IPC 506} & \\
\hline
\multicolumn{10}{|l|}{\hspace{7.5cm} \textbf{Statute Prediction Only}} \\
\hline
\multirow{4}{*}{\textit{LM Based}} & InLegalBERT & 0.76&0.80 &0.83 &0.87 &0.94 &0.91 &0.45 &\textbf{0.82}  \\
\cline{2-10}
 
 &LegalBERT & 0.68 &0.72&0.84 &0.90 &0.93 &0.91 & 0.37&0.80 \\
\cline{2-10}
 & BERT-base &0.60 &0.55 &0.68 &0.81 &0.80 &0.86 &0.06 &0.67 \\
\cline{2-10}
 & Longformer &0.57 &0.59 &0.73 &0.68 &0.83 &0.86 &0.02 & 0.67\\
 \cline{2-10}
 & RoBERTa &0.64 &0.41&0.64&0.75 &0.86 &0.85 &0.10& 0.64\\
 
\hline
\hline
\multirow{4}{*}{\textit{Zero-shot}} & DeepSeek & 0.68 & 0.65&0.75 &0.83 &0.75 &0.73 &0.26 &\textbf{0.64} \\
\cline{2-10}
 & Llama-3.1-70B & 0.47& 0.27& 0.66&0.47 &0.58 &0.55 &0.25 &0.47 \\
\cline{2-10}
 & Claude &0.60 &0.45 &0.71 &0.66 &0.73 &0.71 &0.39 &0.61 \\
\cline{2-10}
 & GPT-4 & 0.55&0.37 &0.68 &0.59 &0.69 &0.66 &0.35 &0.56 \\
\hline
\hline
\multirow{4}{*}{\textit{Few-shot}} & DeepSeek & 0.87 & 0.72 &0.75 &0.85 &0.82 &0.80 &0.43 &\textbf{0.71} \\
\cline{2-10}
 &Llama-3.1-70B &0.66 & 0.43& 0.72&0.83 &0.83 &0.78 &0.37 &0.64 \\
\cline{2-10}
 & Claude & 0.62& 0.52&0.74 &0.74 &0.71 &0.74 &0.41 & 0.64\\
\cline{2-10}
 & GPT-4 & 0.61 &0.51 &0.73 &0.73 &0.78 &0.77 &0.46 & 0.66\\
 
\hline
\hline
\multirow{4}{*}{\textit{TOT}} & DeepSeek &0.21& 0.08&0.12 &0.34 &0.05 &0.21 &0.24&0.17 \\
\cline{2-10}
 & Llama-3.1-70B &0.25& 0.01& 0.09 &0.30 &0.01 &0.24&0.06 &0.13 \\
\cline{2-10}
 & Claude &0.32& 0.02&0.08 &0.74 &0.11 &0.02 &0.03 &0.18 \\
\cline{2-10}
 & GPT-4 &0.05& 0.26& 0.13 &0.36 &0.41 &0.01 &0.27 &\textbf{0.21} \\
 \hline
\hline
\multicolumn{10}{|l|}{\hspace{7 cm} \textbf{Statute Prediction with Explanation}} \\
\hline

\multirow{4}{*}{\textit{COT}} & DeepSeek &0.65 &0.66 &0.74 &0.91 &0.78 &0.88 &0.40 &0.73 \\
\cline{2-10}
 & Llama-3.1-70B &0.50 &0.44 &0.71 &0.71 &0.56 &0.74 &0.33 &0.57 \\
\cline{2-10}
 & Claude & 0.49 &0.58 &0.66 &0.86 &0.82 &0.82 & 0.40&0.66 \\
\cline{2-10}
 & GPT-4 & 0.50&0.61 &0.73 &0.93 &0.93 &0.86 &0.54 &\textbf{0.75} \\
\hline

\end{tabular}
\end{adjustbox}

\label{tab:model_comparison}
\end{table*}
\text

\begin{table*}[t]
\centering
\caption{Comparison of explanation performance across various model combinations for statute prediction using multiple evaluation metrics, including legal expert evaluation scores. The highest value for each metric is highlighted in \textbf{bold}.}
\label{tab:explan_result}
\resizebox{0.8\textwidth}{!}{%
\begin{tabular}{|l|c|c|c|c|c|c|c|}
\hline
\multirow{2}{*}{Models} & \multicolumn{4}{c|}{Lexical Based Evaluation} & \multicolumn{2}{c|}{Semantic Evaluation}&\multicolumn{1}{c|}{Expert Evaluation}\\ \cline{2-8}
 & Rouge-1 & Rouge-2 & Rouge-L & BLEU & BLEURT & BERTScore & Rating Score \\ \hline
\multicolumn{8}{|c|}{Prediction with explanation on PROSLEX} \\ \hline
Llama-3.1-70B & 0.17  & 0.18 & 0.21 & 0.25& 0.58 & 0.65 & 3.16 \\ 
\hline
Deepseek & 0.10  &0.13  &0.15  & 0.21 & 0.42 & 0.46 & 3.83 \\ 
\hline
Claude & 0.10  &0.11  & 0.14 & 0.19 & 0.40 & 0.43  & 3.51 \\ 
\hline
GPT-4 & { \bf 0.20}  & {\bf 0.23} & { \bf 0.31}  & {\bf 0. 28} & {\bf 0.63} & \textbf{0.67} & \textbf{3.91} \\ 
\hline

\end{tabular}%
}

\end{table*}

\section{Experimental Setup}\label{sec:experiment}
\SA{In the previous section, we described the entire data set creation process. For the empirical study, we divide our dataset into three parts: training, validation, and testing. We divided our dataset into training (70\%), validation (10\%), and testing (20\%) sets using stratified random sampling. This configuration provides: a) Sufficient training samples (70\%) for model learning across all statute categories, b) a validation set (10\%) for hyperparameter tuning and early stopping, c) an adequately sized test set (20\%) for robust performance evaluation. In addition, this 70:30 split follows standard practice in legal NLP tasks~\cite{pahilajani2024nlp,joshi2024tur} and provides sufficient data for model training while reserving adequate samples for robust evaluation.
The stratification ensured balanced representation of statute categories across all splits. We used a fixed random seed (seed=42) for reproducibility.}
\para{Language Model Based}
In this approach, we utilized several language models, including LegalBERT~\cite{chalkidis2020legal}, InLegalBERT~\cite{paul2023pre}, Longformer\cite{beltagy2020longformer}, RoBERTa \cite{liu2019roberta}, and BERT-based (Large) \cite{devlin2019bert} as baselines for multilabel classification. For training the model, we used a batch size of 32, the Adam optimizer, and a learning rate of 5e-6. The training was conducted over 20 epochs on the PROSLEX dataset. The remaining hyperparameters were set to their default values as provided by the HuggingFace library. 

\para{Large Language Model Based}
By employing LLMs in prediction, we considered \texttt{two approaches}: one involving identifying \textbf{legal statutes only}, and the other focusing on \textbf{statute identification with explanation}. For this prediction task, we employ several models, including  Llama-3.1-70B\footnote{\url{https://ai.meta.com/blog/meta-llama-3-1/}}, deepseek-r1-distill-llama-70b\footnote{\url{https://www.deepseek.com/en}}, claude-sonnet-4-20250514\footnote{\url{https://www.anthropic.com/claude/sonnet}}, GPT-4\footnote{\url{https://openai.com/}}. The temperature parameter was set to 0 for Claude and GPT, and 0.01 for Llama and DeepSeek. Additionally, we standardized the top-p value to 0.95 across all experiments.

\subsection{Zero-shot}
In this approach, we rely exclusively on the \textit{label descriptions} (as detailed in Section~\ref{sec:statute_detail}) for all labels, without providing any example context, as shown in Table~\ref{tab:prompt_template}.

\subsection{Exemplar selection for few-shot}\label{sec:few_shot}  
In this approach, we adopt the \textit{K-nearest neighbor selection} method to choose an exemplar for few-shot learning. We compute semantic similarity between the \textit{statute description with annotated span of the corresponding statutes for each document from the training set}, and then select examples for each statute based on the highest score.

Let $D_{\text{train}} = \{d_1, d_2, \ldots, d_n\}$ 
denote the set of training documents, and $S = \{s_1, s_2, \ldots, s_m\}$ represent the set of statutes. For each document \(d_i\), let \(a_{ij}\) be the annotated span corresponding to statute \(s_j\), and let \(t_j\) denote the textual description of \(s_j\), as illustrated in Figure~\ref{fig:workflow}. We compute the semantic similarity score between the statute description \( t_j \) and the annotated span \( a_{ij} \) using a similarity function \( \text{$\sigma$}(\cdot, \cdot) \), such as cosine similarity in an embedding space, comparing their [CLS] embeddings obtained from InLegalBERT\footnote{\url{https://huggingface.co/law-ai/InLegalBERT}}, which is trained on Indian legal documents.
The pair \( (s_{j^*}, a_{i^*j}) \) is selected as the representative example for statute \( s_j \). For the empirical study, we conduct experiments using $K$ $\in$ \{2,4,6\} examples per statute as few-shot exemplars.

\subsection{Chain-of-thought}
To elicit a series of intermediate reasoning steps while predicting the statutes, we employ the chain-of-thought approach~\cite{wei2022chain}.
\citet{yu2022legal} employed zero-shot Chain-of-Thought (CoT) prompting for statute prediction, generating explanations prior to prediction; however, this approach underperformed compared to methods that did not utilize explanations. In contrast, \citet{vats-etal-2023-llms} applied few-shot CoT prompting \cite{wei2022chain} on a dataset of 45 documents. In our work, we conducted experiments on a significantly larger dataset (\textbf{1623} documents). Table~\ref{tab:explan_result} and Table~\ref{tab:model_comparison} present the comparative performance of the models in terms of both prediction and explanation.

\subsection{Tree-of-thoughts}
\SA{The Tree-of-Thoughts (ToT) framework has been proposed to improve prompting for complex tasks that demand exploration and look-ahead reasoning~\cite{yao2023tree}. Unlike Chain-of-Thought (CoT) prompting, ToT organizes intermediate reasoning steps---referred to as thoughts---into a tree structure. Each thought constitutes a coherent language sequence that incrementally progresses toward the final solution. This tree-based organization enables language models to reason deliberately by evaluating how well different thoughts contribute to solving the task. Furthermore, ToT combines the model's capability to generate and assess thoughts with search strategies such as breadth-first and depth-first search.}

\SA{In this study, we employ a breadth-first search (BFS) strategy to ensure systematic exploration of the solution space at each reasoning level before proceeding deeper. Our hyperparameters were determined through preliminary experiments on a validation subset: we set a maximum search depth of $d_{\text{max}} = 7$ to balance exploration breadth with computational feasibility, allow at most $c_{\text{max}} = 3$ child nodes per state to maintain diversity while avoiding combinatorial explosion, impose a limit of $n_{\text{max}} = 10$ nodes per level to constrain memory usage, and apply a confidence threshold of 0.4 (representing the minimum normalized probability score for path expansion) to prune low-quality reasoning branches early. We chose BFS over depth-first search because it proved more robust in our pilot studies for tasks requiring comparison of multiple reasoning strategies at similar depths. These parameters result in a maximum of $3^7 = 2{,}187$ possible paths in the worst case, though early pruning typically reduces the explored space to 15--30\% of this theoretical maximum.}

\section{Results and Analysis}
In this section, we discuss the insights derived from our empirical observations. Table~\ref{tab:model_comparison} shows the F1-score for each statute, followed by the macro-F1 score, while Table~\ref{tab:explan_result} compares LLM-generated explanations with expert annotations.

\subsection{Insights from Statute Prediction}
Specifically, Table~\ref{tab:model_comparison} provides a comparative analysis of the performance of different models across the aforementioned sections of the Indian Penal Code (IPC), using both traditional language-based models and Large Language Models (LLMs). For the LLM-based systems, we further explore \textit{in-context learning} (ICL), \textit{Chain-of-Thought} (CoT), and \textit{Tree-of-Thoughts} (ToT) strategies.  InLegalBERT emerges as the best-performing model under the ``Statute Prediction Only'' setting, achieving the highest overall accuracy of 0.82, notably outperforming other language models like LegalBERT and Longformer. Among \textit{zero-shot} setups, DeepSeek performs best with 0.64 macro F1-score, though still significantly behind the fine-tuned LM-based models. In the \textit{few-shot} setting, DeepSeek again leads with 0.71 macro F1-score, demonstrating better generalization with limited examples, while GPT-4 and Claude show more moderate performance. 

Another observation is that LLMs under the \textit{ToT} mechanism struggle with accurate statute prediction. When generating multiple reasoning branches, the model may interpret the same statutory language differently across branches, resulting in inconsistent or incorrect predictions. In contrast, \textit{CoT} follows a single reasoning chain. This focused strategy allows LLMs to concentrate computational resources on one line of reasoning rather than dispersing attention across multiple branches, leading to more accurate statute identification. Consequently, GPT-4 achieves the highest macro F1-score of 0.75, demonstrating strong reasoning capabilities, particularly when explanations are required.

Overall, while domain-specific models like InLegalBERT dominate in direct statute prediction, large foundational models like GPT-4 excel when interpretability and reasoning are involved.

\begin{table*}[!ht]
\caption{Shows that LLM-generated explanations and their associated IPC sections are verified by a senior legal expert (discussed in Section~\ref{sec:insights_reasoning}).}
\centering
\small
\begin{tabular}{p{2.2cm} p{9.5cm} p{2.2cm}}
\hline
\hline
\textbf{Inferred Section} & \textbf{Generated Explanation} & \textbf{Human Score} \\
\hline
\hline
 \hspace{0.5cm} IPC 302 & Appellant no. 2 gave a blow with an axe on the neck of the deceased as a result of which he fell down and soon thereafter died. & \hspace{1cm} 4 \\

 \hspace{0.5cm} IPC 302 & The deceased had come to the house of PW 1, heard sound of falling of a cycle. He therefore looked into that direction and saw that the appellant was giving blows to his friend with a sword. & \hspace{1cm} 4 \\

 \hspace{0.5cm} IPC 420 & Kanhayalal Sharma and his son Prakash Kanhayalal Sharma are being investigated for having issued bogus and false Degree certificates to a large number of students. & \hspace{1cm} 4 \\

 \hspace{0.5cm} IPC 420 & The appellant allotted premises to various persons under his signature, issued rent receipts so that the said persons could claim possession of the tenements, though in fact the tenements were vacant. & \hspace{1cm} 4 \\

 \hspace{0.5cm} IPC 147 & Companions of Kallu came to the factory and murdered Ashok Kumar. & \hspace{1cm} 0 \\

 \hspace{0.5cm} IPC 147 & Allegations against them were that on September 17, 1982, they committed rioting. & \hspace{1cm} 2 \\

 \hspace{0.5cm} IPC 506 & The appellant took out a revolver and threatened the complainant with death unless he withdrew the complaint. & \hspace{1cm} 2\\

 \hspace{0.5cm} IPC 376 & The doctor opined that Asha was possibly raped. & \hspace{1cm} 4\\

 \hspace{0.5cm} IPC 201 & The claimants also claimed 12\% additional compensation under Section 23(1A) of the L.A. Act, which the Court below had not granted. & \hspace{1cm} 4 \\

 \hspace{0.5cm} IPC 201 & It was alleged that the chopped portion of the head was handed over to the accused Abdullah. & \hspace{1cm} 0 \\
\hline
\hline
\end{tabular}

\label{tab:ipc_human_eval}
\end{table*}

\subsection{Insights of LLM Reasoning}\label{sec:insights_reasoning}
To evaluate the LLMs generated rationale along with predicted statutes, we employ a comprehensive evaluation strategy that integrates both quantitative and qualitative analyses. This multifaceted approach ensures a robust assessment of our model's performance for the explanation tasks.

\begin{enumerate}
    \item \textbf{Lexical Based Evaluation:} We employed lexical similarity metrics, including ROUGE~\cite{lin2004rouge} (ROUGE-1, ROUGE-2, ROUGE-L), and BLEU~\cite{papineni2002bleu} to evaluate the generated explanations. These metrics measure the overlap and sequence of words between the model outputs and reference texts, offering insight into the lexical accuracy of the explanations.
    \item \textbf{Semantic Similarity Based Evaluation:}
    To capture the semantic quality of the generated explanations, we used BERTScore~\cite{zhang2019bertscore}, BLEURT~\cite{sellam2020bleurt}, which evaluate the semantic similarity between the generated outputs and the reference explanations. Therefore, these metrics provide deeper insight into the model’s ability to produce meaningful and contextually appropriate explanations.
    \item \textbf{Expert Evaluation:}
    Evaluating generative models in the statute identification task with explanation requires domain-specific expertise.
    We shared LLM generated predicted statutes along with their explanations to legal experts and asked them to rate the alignment between explanations and predicted statutes on a scale from 0 to 4:

    \begin{itemize}
    \item \textbf{4:} Fully aligned - explanation directly and completely supports the predicted statute.
    \item \textbf{3:} Mostly aligned - explanation supports the statute with minor gaps or tangential points.
    \item \textbf{2:} Partially aligned - explanation has some relevant points but also contains misaligned or irrelevant content.
    \item \textbf{1:} Minimally aligned - explanation barely relates to the predicted statute.
    \item \textbf{0:} Not aligned - explanation is irrelevant or contradicts the statute.
\end{itemize}

\end{enumerate}
Table~\ref{tab:explan_result} presents a comparative analysis of explanation quality across various models such as LLaMA- 3.1-70B, DeepSeek, Claude, and GPT-4, employing both lexical-based (ROUGE-1, ROUGE-2, ROUGE-L, BLEU) and semantic-based (METEOR, BERTScore) evaluation metrics. Overall, GPT-4 consistently outperforms the other models across all metrics, achieving the highest scores in ROUGE-1 (0.20), ROUGE-2 (0.23), ROUGE-L (0.31), BLEU (0.28), BLEURT (0.63), and BERTScore (0.67), indicating that its explanations are both textually aligned and semantically richer compared to other models. LLaMA 3.1 ranks second, showing relatively strong lexical overlap (e.g., ROUGE-2: 0.18) and decent semantic quality (BLEURT: 0.58, BERTScore: 0.65), suggesting it generates explanations that are closer to reference texts than DeepSeek or Claude. DeepSeek and Claude lag behind significantly on all metrics, with Claude performing the lowest semantically (BERTScore: 0.43), indicating less coherent or relevant explanations. This analysis confirms that GPT-4 is the most capable model for generating legally meaningful and well-aligned statute explanations, with LLaMA-3.1-70B being a strong secondary contender. 
Among the evaluated models, GPT-4 achieves the highest expert rating with an average score of 3.91, indicating the most consistently high-quality explanations as assessed by domain experts. Deepseek follows with a strong expert score of 3.83, suggesting that its explanations are generally well-aligned with legal expectations. Claude attains an average rating of 3.51, reflecting competent but comparatively less consistent performance.

In contrast, Llama-3.1-70B receives the lowest expert rating (3.16), indicating greater variability in explanation quality and more frequent deviations from expert standards. Overall, the expert evaluation results highlight clear performance differences across models, with GPT-4 and Deepseek demonstrating superior alignment with expert legal reasoning criteria.

\SA{Table~\ref{tab:rating_score_distribution} reports the distribution of expert-assigned rating scores for explanations generated by different generative models under the PROSLEX setting. Scores range from 0 to 4, with higher values indicating better explanation quality.}

\SA{Across all models, the majority of explanations receive the highest rating (score 4), indicating consistently strong performance. GPT-4 achieves the highest concentration of top-rated explanations, with 47 instances rated 4 and no explanations receiving scores 0 or 1. Deepseek and Claude show similarly strong performance, with 46 and 43 explanations, respectively, receiving the maximum score, although Deepseek exhibits a small number of lower-rated cases.}

\SA{In comparison, Llama-3.1-70B demonstrates greater variability, with 38 explanations rated 4 and 10 explanations receiving a score of 0 (including cases where the model failed to generate any explanation), indicating low-quality outputs. Ratings of 1 and 2 are rare across all models, suggesting that explanations are generally evaluated as either high quality or clearly inadequate rather than marginal.}

\SA{Overall, the distribution indicates that all evaluated models are capable of producing high-quality explanations, with GPT-4 showing the most consistent performance in terms of expert ratings. Below, we provide an error analysis to illustrate common failure modes, followed by examples of expert-evaluated explanations.}
\begin{figure*}[t]
\centering
\begin{tcolorbox}[title={An example of Misaligned Reasoning Despite Correct Statute Prediction}]
\footnotesize
  ... The appellant was bewildered and motionless for some time. The appellant apprehended that Dr. Bhattacharya would create a situation which may adversely affect appellant's health. There was heated exchange of words which resulted in commotion. \textcolor{red}{\textbf{There was scuffle on the arrival of outsiders and two sons of Parul Jana out of hospital premises.}} The appellant immediately contacted the General Manager (Projects) and requested him to help to control the situation. When the General Manager reached the hospital, the appellant explained the situation to him. The General Manager also met Dr. Bhattacharya to get true and correct facts as to how the incident had happened. The General Manager then advised the appellant to go back. Immediately, the appellant left the hospital. In the entire incident, asserted the appellant, save and except accompanying sons of Parul Jana, he did nothing. He was not involved in the incident in any manner whatsoever. It was the Chief Medical Officer, who alone was responsible for the entire unfortunate situation. He also inflicted injuries on two sons of Parul Jana. Dr. Bhattacharya, however, cooked up a false case against the appellant alleging that the appellant had assaulted and injured him. On 6th May, 1999, i.e. on the next day, the Chief Medical Officer, \textcolor{blue}{ \textbf{Dr. Bhattacharya reported to the management that at the late night hours of 5th May, 1999, the appellant led by a bunch of hooligans had visited the hospital, assaulted him,}} i.e. Dr. Bhattacharya and abused and threatened other officers....
\end{tcolorbox}
\caption{The red-colored text span represents the portion labeled by the GPT-4 model in the \textit{Chain-of-Thought} setup, where the predicted statute is \textit{Indian Penal Code 147}. The blue-colored span corresponds to the annotation provided by legal experts, with the same predicted statute, \textit{Indian Penal Code 147}. Thus, although GPT-4 correctly identified the statute, its reasoning was incorrect. }
\label{fig:bad_analysis}
\end{figure*}
\begin{table}[t]
\centering
\caption{Distribution of rating scores for different generative models is assigned by the legal experts for each explanation.}
\Large
\begin{tabular}{lccccc}
\hline
\hline
\textbf{Generative Models} & \multicolumn{5}{c}{\textbf{Rating Score}} \\
\cline{2-6}
 & \textbf{0} & \textbf{1} & \textbf{2} & \textbf{3} & \textbf{4}  \\
\hline
\hline
\multicolumn{6}{l}{\textbf{PROSLEX}} \\
\hline
Llama-3.1-70B & 10 & 0 & 0 & 1 & 38  \\
Deepseek & 1 & 0 & 2 & 0 & 46  \\
Claude & 6& 0 & 0 & 0 & 43 \\
GPT-4 & 0 & 0 & 2 & 0 & 47 \\
\hline
\end{tabular}

\label{tab:rating_score_distribution}
\end{table}

\subsection{Error Analysis}\label{sec:error_analysis}
Figure~\ref{fig:bad_analysis} illustrates this challenge. 
The model (GPT-4) correctly predicts the applicable statute (IPC 147) but provides misaligned reasoning. The \emph{model-selected span} (red) highlights a fight, which does not sufficiently establish the statutory elements of rioting, whereas the \emph{expert-annotated span} (blue) explicitly describes the appellant leading a group that assaulted and threatened hospital staff, satisfying the legal requirements of IPC 147. The model relies on surface-level indicators of physical altercation rather than legally salient facts, demonstrating that correct statute prediction does not necessarily imply correct legal reasoning.

Table~\ref{tab:ipc_human_eval} further reveals this variability: while LLMs can generate persuasive and contextually rich legal explanations for some IPC sections, their outputs are not uniformly reliable across all cases. This reinforces the need for human-in-the-loop validation in legal AI systems.

Overall, this error analysis underscores the importance of fine-grained, expert-aligned supervision of reasoning spans and evaluation frameworks that assess both predictive correctness and the legal validity of supporting explanations.

\section{Conclusion and Future Work}

This work introduces \texttt{PROSLEX}, the largest and most comprehensive dataset constructed to date for the challenging task of legal statute prediction and explanation. By providing an unprecedented scale of annotated legal cases paired with relevant statutory provisions and human-generated explanations, \texttt{PROSLEX} represents a significant milestone in advancing computational approaches to legal reasoning. The dataset addresses a critical gap in the legal AI landscape, where previous resources have been limited in scope, coverage, or the depth of explanatory annotations required to train and evaluate sophisticated models.

\section*{Acknowledgments}
We acknowledge the Start-Up grant provided by IISER Kolkata to Dr. Kripabandhu Ghosh for covering the annotation cost. Generative AI tools were not used for core idea generation or experimental design. Its use was limited to minor writing and formatting.

\bibliographystyle{ACM-Reference-Format}
\bibliography{bibliography}

\appendix

\end{document}